\documentclass[11pt]{article}

\usepackage[preprint]{acl}

\usepackage{multirow}
\usepackage{times}
\usepackage{latexsym}
\usepackage{amsmath, amssymb}
\usepackage{algorithm}
\usepackage{algorithmic}

\usepackage[T1]{fontenc}

\usepackage[utf8]{inputenc}

\usepackage{microtype}

\usepackage{inconsolata}

\usepackage{graphicx}
\usepackage{amsmath}
\usepackage{amssymb}
\usepackage{amsfonts}
\usepackage{booktabs}
\usepackage{geometry}
\usepackage[table]{xcolor}
\usepackage{booktabs}

\usepackage[most]{tcolorbox}
\definecolor{darkgrey}{rgb}{0.53,0.53,0.53}
\definecolor{mygrey}{rgb}{0.9,0.9,0.9}
\definecolor{purple}{RGB}{230, 227, 254}
\definecolor{lightgreen}{RGB}{238, 252, 241}
\definecolor{lightred}{RGB}{231, 187, 187}
\definecolor{darkred}{RGB}{198, 129, 129}

\definecolor{tabhighlight}{HTML}{e5e5e5}
\definecolor{someorange}{rgb}{0.773,0.353,0.067}
\definecolor{someblue}{rgb}{0.27, 0.35, 0.760}

\title{C$^2$DLM: Causal Concept-Guided Diffusion Large Language Models}

\author{
 \textbf{Kairong Han\textsuperscript{1}},
 \textbf{Nuanqiao Shan\textsuperscript{1}},
 \textbf{Ziyu Zhao\textsuperscript{1}},
 \textbf{Zijing Hu\textsuperscript{1}},
 \textbf{Xinpeng Dong\textsuperscript{1}},\\
 \textbf{Junjian Ye\textsuperscript{2}},
 \textbf{Lujia Pan\textsuperscript{2}},
 \textbf{Fei Wu\textsuperscript{1}},
 \textbf{Kun Kuang\textsuperscript{1\dag}}\\
 \textsuperscript{1}College of Computer Science and Technology, Zhejiang University,\\
 \textsuperscript{2}Noah's Ark Lab, Huawei Technologies,\\
}

\begin{document}
\maketitle
\begin{abstract}
Autoregressive (AR) language models and Diffusion Language Models (DLMs) constitute the two principal paradigms of large language models.  However, both paradigms suffer from insufficient reasoning capabilities. Human reasoning inherently relies on causal knowledge and thought, which are reflected in natural language. But in the AR paradigm, language is modeled as next token prediction (a strictly left-to-right, token-by-token order), whereas natural language itself exhibits more flexible causal structures. In the DLM paradigm, the attention mechanism is fully connected, which entirely disregards causal order. To fill this gap, we propose a \underline{\textbf{C}}ausal \underline{\textbf{C}}oncept-Guided \underline{\textbf{D}}iffusion \underline{\textbf{L}}anguage \underline{\textbf{M}}odel (C$^2$DLM). Starting from DLM's fully connected attention, C$^2$DLM first obtains a concept-level causal graph from the teacher model, and then explicitly guides attention to learn causal relationships between concepts. By focusing on causal relationships and avoiding interference from difficult subgoals involving causal inversion, C$^2$DLM improves 12\% with a about 3.2× training speedup in the COT-OrderPerturb task, and achieves an average gain of 1.31\% across six downstream reasoning tasks. More details in the repository ~\href{https://github.com/Kairong-Han/C-2-DLM}{here}.
\end{abstract}

\section{Introduction}

In recent years, the development of large language models (LLMs) \cite{zhao2023survey,liu2024deepseek,team2023gemini} has led to two dominant paradigms: autoregressive (AR) LLMs and diffusion language models (DLMs) \cite{li2025survey,nie2025large}. In the AR paradigm, a causal mask \cite{vaswani2017attention} constrains the model with a lower-triangular matrix to predict the next token based on preceding tokens. In contrast, the DLM paradigm employs fully connected attention to guide the model in globally modeling data from coarse to fine \cite{ye2024beyond}.
\begin{figure}[t]
    \centering
\includegraphics[width=1\linewidth]{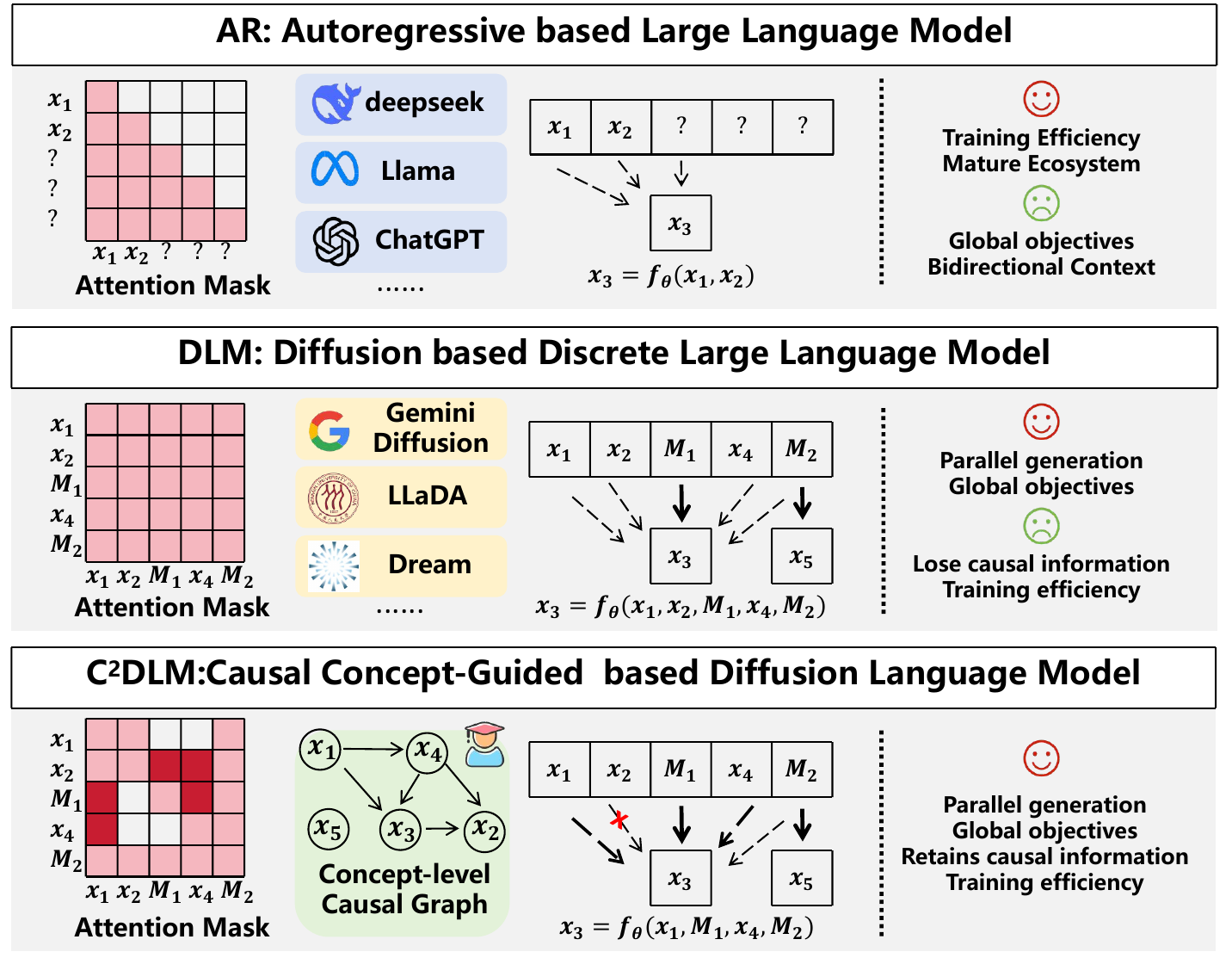}
\caption{Difference between AR, DLM, and C$^2$DLM. AR models struggle to capture global information, and linguistic flexibility is not bound to a strict left-to-right, token-by-token causal order. DLMs discard causal priors entirely. The C$^2$DLM explicitly guides the model to learn causal relations between concepts, capturing the underlying causal priors of natural language generation.}
    \label{fig:intro}

\end{figure}

However, both AR and DLMs suffer from insufficient reasoning capabilities, such as frequent hallucinations \cite{huang2025survey} and unreliable reasoning chains \cite{lanham2023measuring,yehudai2025survey}, imposing fundamental limitations on tasks that require reasoning and posing significant challenges in real-world deployment \cite{pan2025multiagent,acharya2025agentic,wu2024causalitylargelanguagemodels}. Specifically, AR models exhibit limitations in complex reasoning, long-term planning, and maintaining global coherence \cite{ye2024beyond,bubeck2023sparks,kambhampati2024llms,zevcevic2023causal}. DLMs, as strong competitors to AR models, reduce training efficiency and hinder the effective scaling of reasoning depth, thereby limiting their potential for complex tasks.

Human reasoning inherently relies on causal knowledge and thought. From a natural language perspective, it is inherently flexible rather than strictly left-to-right and token-by-token causal structures. However, AR generation enforces unidirectional information flow, resulting in local greediness and a limited understanding of global objectives and long-term structure. In contrast, DLMs discard the causal order between tokens, often producing final answers before the intermediate reasoning steps that CoT methods would generate \cite{wang2025time}. Their training further involves numerous difficult sub-tasks (e.g., predicting masked cause variables from outcomes under random masking) \cite{kim2025train}, which reduces training efficiency and constrains the scalable development of reasoning depth.

To address the above fundamental problems, we hypothesize that \textit{these limitations stem from a misalignment between the attention mechanism’s modeling priors on natural language and the causal priors underlying natural language}. Therefore, we aim to guide the model to capture the underlying causal priors of the natural language generation process, rather than superficial correlations. Inspired by this, we propose \underline{\textbf{C}}ausal \underline{\textbf{C}}oncept-Guided \underline{\textbf{D}}iffusion \underline{\textbf{L}}anguage \underline{\textbf{M}}odel (C$^2$DLM) paradigm, as shown in Figure \ref{fig:intro}.

The C$^2$DLM extends DLMs by two key steps: (1) concept-level causal meta-knowledge extraction, and (2) causal alignment via the \textit{V-aware Re-attention} mechanism. In the first step, to obtain concept-level causal graphs at low cost, an automated workflow leverages the in-context learning (ICL) \cite{dong2022survey} capabilities of teacher LLMs to extract concept-level meta-knowledge. In the second step, we propose the V-aware Re-attention mechanism to align the attention map weighted by the L2-norm of the Value matrix with the underlying causal priors of the natural language generation process generated by step one.

To compare AR, DLM, and C$^2$DLM systematically, we design the COT-OrderPerturb dataset to quantify the impact of priors. AR models are sensitive to concept order, whereas DLMs are more robust but limited by efficiency and performance bottlenecks. Building on DLMs, C$^2$DLM achieves a 12\% higher performance and 3.2× faster training. On downstream tasks with explicit causal priors, C$^2$DLM yields average improvements of 7.43\% on STG \cite{han2025cat} and 10.84\% on Sudoku (training set size 200). Across six reasoning-related datasets, it delivers an average gain of 1.31\%, while causal prior extraction with GLM-4.5 costs only \$0.46 per million tokens. Our contributions can be summarized as follows:

\begin{itemize}
    \item We propose C$^2$DLM, a new paradigm distinct from AR and DLM. It enhances reasoning ability by guiding attention through causal knowledge between concepts to achieve causal alignment.

    \item The C$^2$DLM achieves 12\% improvement and a 3.2× acceleration of training efficiency in the COT-OrderPerturb tasks, 7.43\% on the STG dataset, and 1.31\% across six reasoning-related downstream datasets on average.

    \item We reveal the risk of misalignment between attention mechanisms and the causal priors underlying natural language, which shows the potential of combining causality into language models.
\end{itemize}

\section{Preliminaries and Related Works}

\subsection{Diffusion Large Language Model}\label{sec:2.1}
 Recently, researchers have adapted the diffusion paradigm \cite{yang2023diffusion,cao2024survey} to discrete text data, proposing DLM  \cite{nie2025large,ye2025dream,ye2024beyond,austin2021structured}, which achieve competitive performance compared to  AR models. DLMs employ a bidirectional attention mechanism and leverage the Negative Evidence Lower Bound to provide an upper bound on the negative log-likelihood of the training data, thereby modeling the distribution of language. LLaDA \cite{nie2025large} first demonstrated the effectiveness of diffusion at the 8B parameters, using the following supervised fine-tuning (SFT) loss $\mathcal{L_{DLM}}$:

$$
 - \mathbb{E}_{t, p_0, r_0, r_t} \Bigg[ \sum_{i=1}^{L'} \mathbf{1}[r_i^t = M] \cdot \log p_\theta(r_i^0 \mid p_0, r_t) \Bigg],
$$

where $r_t$ denotes the noised sequence appended to the prompt $p_0$. Recent work on DLMs has mainly focused on improving DLM by reinforcement learning (RL) \cite{kaelbling1996reinforcement}, such as d1 \cite{zhao2025d1},wd1 \cite{tang2025wd1}, BranchGRPO \cite{li2025branchgrpo}, and et al. \cite{zhao2025inpainting,zhu2025llada}. Another line of work focuses on speeding up DLM inference time, such as Fast-dllm \cite{wu2025fast}, SlowFast \cite{wei2025accelerating}, and et al. \cite{wang2025diffusion,hu2025accelerating}.

However, C$^2$DLM focuses on causal alignment of the attention mechanism in the SFT stage.

\subsection{Combining Causality and Attention Mechanism}
Transformer \cite{vaswani2017attention} proposes the multi-head attention mechanism, which models dependencies between tokens:
$$\textbf{A}_i^{attn} =  \text{softmax}\left(\frac{\mathbf{Q}_i \cdot \mathbf{K}_i^\top}{\sqrt{ d_k}}\right) \cdot \mathbf{V}_i,
$$
where for each head $i$ in multi-head attention, $\mathbf{Q}_i,\mathbf{K}_i \in \mathbb{R}^{n \times d_k}$,$\mathbf{V}_i \in \mathbb{R}^{n\times d_v}$.

Although attention and causal \cite{pearl2009causality} graphs are correlated in the covariance structure \cite{rohekar2023causal}, the attention sink \cite{sun2024massive,xiao2023efficient,gu2024attention} phenomenon reveals outliers in the attention distribution, reducing interpretability \cite{kobayashi-etal-2020-attention}. To correct and denoise attention, some studies leverage causal backdoor mechanisms for debiasing in text \cite{Wu_Liu_Zhao_Lu_Zhang_Sun_Wu_Kuang_2024} and vision \cite{yang2021causal}. Recent work proposed the Re-attention mechanism \cite{han2025cat} to inject causal knowledge into a student model.

Our C$^2$DLM is the new paradigm to investigate attention mechanisms in DLMs, interpreting the importance of guiding the model to align with data generation priors.

\subsection{The Limitations of Attention in AR and DLM}

The limitations of AR and DLM models stem from the structural priors imposed by their attention mechanisms.
The AR model, with its lower-triangular attention matrix and modeling objective $P(x) = \prod_i p_{\theta}(x_i | x_{<i})$, is unable to handle situations in natural language where the outcome precedes the cause. On the other hand, DLM can be regarded as an any-order AR model \cite{arriola2025block}. DLM adopts a fully connected structure that can model arbitrary dependencies: $P(x_i) = p_{\theta}(x_i \mid x_{\neq i})$. The absence of causal constraints causes key causal signals in each step of the COT process to be diluted by redundant information from both past and future contexts. This makes it difficult for the model to perform stable and effective reasoning over COT.

\section{Method}

\begin{figure*}
    \centering
    \includegraphics[width=1\linewidth]{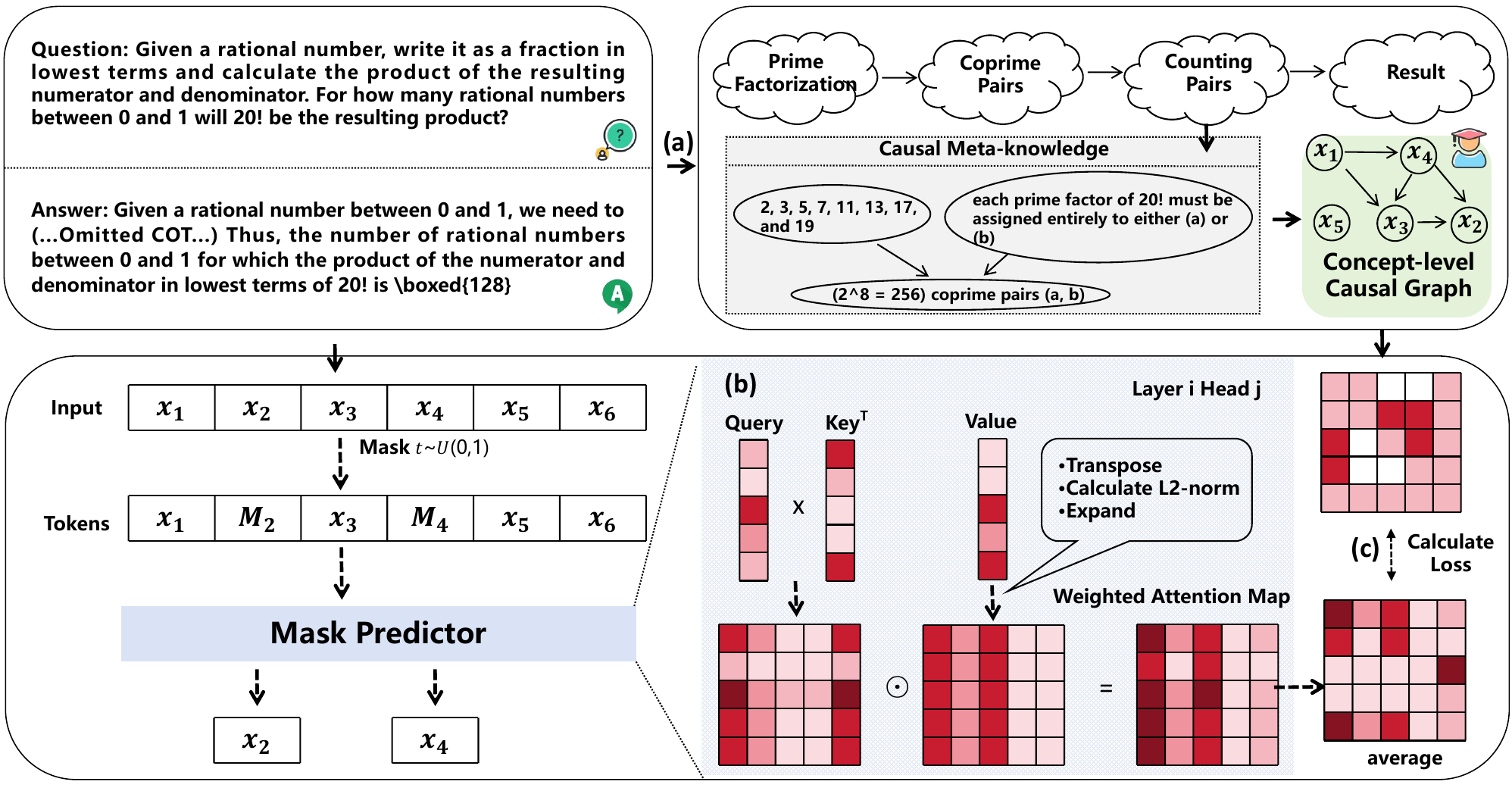}
    \caption{(a) Leveraging the contextual learning capability of a strong model, the causal teacher model uses prompts to automatically extract concept-level information from  CoTs and generates causal meta-knowledge links between concepts as supervisory signals. (b) During training, for the internal attention map obtained from CoTs, the V-aware Re-attention mechanism weights the attention maps by the norms of the corresponding Value matrix. (c) The tokenizer maps the textual supervisory signals from step (a) to the weighted attention maps, and a loss-based intervention is applied to guide the C$^2$DLM’s decision-making process.}
    \label{fig:method}
\end{figure*}

As shown in Figure \ref{fig:method}, C$^2$DLM consists of two main steps: (1) concept-level causal meta-knowledge extraction, and (2) causal alignment via the V-aware Re-attention mechanism.

\subsection{Concept-level Causal Meta-knowledge Extraction}\label{sec:Concept-level}

Humans, when confronted with downstream tasks, analyze relationships among conceptual entities, perform reasoning, and integrate contextual information to verbalize their thought process in natural language. Inspired by this, we extract and construct concept-level reasoning graphs for tasks described in natural language. Each concept encapsulates the core information necessary for reasoning and reflects human causal logic, thereby encoding the true and flexible priors underlying the data. Therefore, the generating function of language should be consistent with human’s prior understanding of causal concepts.

To reduce the cost of generating such priors, we design an automated workflow. The teacher model first extracts a set of concepts  $\mathcal{C} = \{c_1, c_2, \ldots, c_n\} $ from the reasoning steps, and denotes the remaining text as context  $\mathcal{T}$. Each concept $c \in \mathcal{C}$ represents a semantically complete entity or sentence. The teacher then constructs a reasoning graph over $ \mathcal{C}$, capturing causal dependencies between concepts.

Unlike prior work \cite{han2025cat}, the graph is not restricted to pairwise causal forms such as $c_A$ causes $c_B$. For a given concept $c_A$, information from $c_B$ that is unnecessary for generation is pruned, preventing reverse dependencies. For example, as shown in Figure \ref{fig:method}(a), the teacher decomposes the problem into four steps and identifies causal meta-knowledge within each step. Conditions like “2, 3, 5, 7, 11, 13, 17, 19” and “each prime factor of 20! must be assigned entirely to a or b” together imply 256 valid (a, b) pairs, though this fact is not required when generating those conditions. For certain tasks, we further introduce a rule-based, semi-autoregressive supervisory signal that decomposes the reasoning chain into coarser-grained steps $s \in \mathcal{S}$ based on inter-rule inference. This mechanism prunes irrelevant information from earlier steps, improving the efficiency of subsequent reasoning. Prompt details are provided in Appendix \ref{APP:Prompt}.

Based on the above constraints, we define the prior supervision mask as:
\[
M_{i,j} =
\begin{cases}
1,  & c_i, c_j \in \mathcal{C},~ c_i \rightarrow c_j, \\[2mm]
0,  & c_i \in \mathcal{T} \text{ or } c_j \in \mathcal{T}, \\[2mm]
-1, & c_i, c_j \in \mathcal{C},~ c_j \rightarrow c_i \text{ or } s_i>s_j.
\end{cases}
\]
where $c$ and $s$ are concepts and steps indexed by token $i$ and $j$ separately.

\subsection{Causal Alignment via the V-aware Re-attention Mechanism}

To align the model’s decision dependencies with the underlying causal mechanisms of natural language and eliminate instability caused by outliers in the attention map, we propose the \textit{V-aware Re-attention} mechanism. With respect to the supervisory mask introduced in the previous section, we define the index sets as
\[
I_k = \{ j \mid M_{i,j} = k \}, \quad k \in \{-1,0,1\}.
\]

Because the attention map can be distorted by the attention sink phenomenon, its raw values may fail to accurately reflect token-level interactions. To address this, we incorporate the L2-norm of the Value matrix as weighting information. Let $A^{(h)} \in \mathbb{R}^{T_q \times T_k}$ denote the attention map of the $h$-th head, and $V^{(h)} \in \mathbb{R}^{T_q \times d_h}$ the corresponding Value matrix. We use the L2 norm of  the Value matrix as weighted information  \cite{kobayashi-etal-2020-attention}:
\[
\| V_i^{(h)} \|_2 = \sqrt{\sum_{d=1}^{d_h} \big(V_{i,d}^{(h)}\big)^2}, \quad i = 1,\dots,T_q.
\]
The weighted attention map is then
\[
\widetilde{A}_{i,j}^{(h)} = A_{i,j}^{(h)} \cdot \| V_i^{(h)} \|_2,
\quad \forall i \in [1,T_q], \; j \in [1,T_k],
\]
and averaging across $n_h$ heads yields
\[
\widetilde{A}_{i,j} = \frac{1}{n_h} \sum_{h=1}^{n_h} \widetilde{A}_{i,j}^{(h)}.
\]

Based on $\widetilde{A}$, we compute the average attention values for encouraged and neutral sets as
\[
\bar{A}_1 = \frac{1}{|I_1|} \sum_{j \in I_1} \widetilde{A}_{i,j},
\qquad
\bar{A}_0 = \frac{1}{|I_0|} \sum_{j \in I_0} \widetilde{A}_{i,j}.
\]
The ratio loss for the $i$-th row is
\[
\mathcal{L}_{ratio}(i) =
\begin{cases}
- \dfrac{\bar{A}_1}{\bar{A}_1 + \bar{A}_0}, & \dfrac{\bar{A}_1}{\bar{A}_0} < \alpha, \\\\[2mm]
 0, & \text{otherwise},
\end{cases}
\]
where $\alpha > 0$ enforces a minimum ratio between encouraged and neutral attentions. In addition, for masked entries with $M_{i,j} = -1$, we penalize the squared weighted attention values:
\[
\mathcal{L}_{neg}(i) = \lambda \sum_{j \in I_{-1}} \widetilde{A}_{i,j}^2,
\]
with $\lambda > 0$ controlling the penalty strength. The total loss for the $i$-th row is then
\[
\mathcal{L}_{row}(i) = \mathcal{L}_{neg}(i)+\mathcal{L}_{ratio}(i).
\]
For the $\mathcal{J}_r$ valid rows that have supervisory signals, we apply weighting and combine them with the DLM downstream SFT loss (as described in Section~\ref{sec:2.1}) to obtain the final training loss:
$$
\mathcal{L}_{total} = \mathcal{L}_{DLM} +  \frac{\gamma}{|\mathcal{J}_r|}\sum_{i\in \mathcal{J}_r}\mathcal{L}_{row}(i),
$$
where $\gamma > 0$ is a balancing coefficient controlling the relative strength of the proposed constraint loss.

Because we directly intervene on the weights of the attention matrix, we introduce a smoothing mechanism to stabilize training. Inspired by learning rate scheduling, we define a $\gamma$-parameter scheduler $\mathcal{S}_\gamma$ that modulates $\gamma$ over training steps. Specifically, for the initial set of steps, $\gamma$ increases linearly from $\gamma_{\min}$ to $\gamma_{\max}$, and for the subsequent steps, it decreases linearly back to $\gamma_{\min}$. Formally, this can be expressed as:

$$
\gamma_t =
\begin{cases}
\gamma_{\min} + \frac{t}{T_1} (\gamma_{\max} - \gamma_{\min}), &  t \in[0,T_1] \\
\gamma_{\max} - \frac{t - T_1}{T_2-T_1} (\gamma_{\max} - \gamma_{\min}), & t \in[T_1,T_2],
\end{cases}
$$
where t denotes the current training step, $T_1$ is the number of warm-up steps, and $T_2-T_1$ is the number of cool-down steps.

\section{Experimental Results}
\subsection{Experimental Setup}
\textbf{Datasets.} We first constructed a synthetic dataset, COT-OrderPerturb, to examine how the ordering of concepts within COT influences AR and DLM. We then employed the Sudoku\footnote{https://github.com/Black-Phoenix/4x4-Sudoku-Dataset} and STG \cite{han2025cat} datasets to assess how models benefit when downstream tasks exhibit explicit causal structures. Finally, we evaluated broader reasoning-related downstream tasks, including MATH500 \cite{lightman2023let}, GSM8K \cite{cobbe2021gsm8k}, GPQA \cite{rein2024gpqa}, ARC\_C \cite{allenai:arc}, SAT \cite{zhong2023agieval}, and MMLU\_STEM \cite{hendryckstest2021,hendrycks2021ethics}.

\textbf{Baselines and hyperparameters.} We adopt LLaDA-8B-Instruct\footnote{https://github.com/ML-GSAI/LLaDA/} as the primary experimental model and apply LoRA \cite{hu2022lora} for fine-tuning, with SFT serving as the DLM main baseline. During training and evaluation, all hyperparameters are consistent except for the loss introduced by C$^2$DLM. In addition, we include the following commonly used AR models for comparison:  Llama-3.1-8B, Llama-3.2-1B \cite{dubey2024llama}, Qwen-2.5-1.5B, and Qwen3-8B \cite{yang2025qwen3}. For the $\gamma$-parameter scheduler, we set $T_1 = 0.1*T_2$. For the $\lambda$-parameter, we set 100 for the COT-OrderPerturb and 10 for all other tasks. The learning rate is uniformly fixed at $2\times e^{-5}$, and the LoRA rank is set to 128. For STG and Sudoku, we set $\alpha$ as 5, and for other tasks, $\alpha$ is 3. Unless otherwise specified, the block length during the test defaults to 32. More detailed hyperparameter configurations are provided in Appendix \ref{APP:Hyper}.

\subsection{Quantifying the Impact of Priors}

\begin{figure}[t]
    \centering
    \includegraphics[width=1\linewidth]{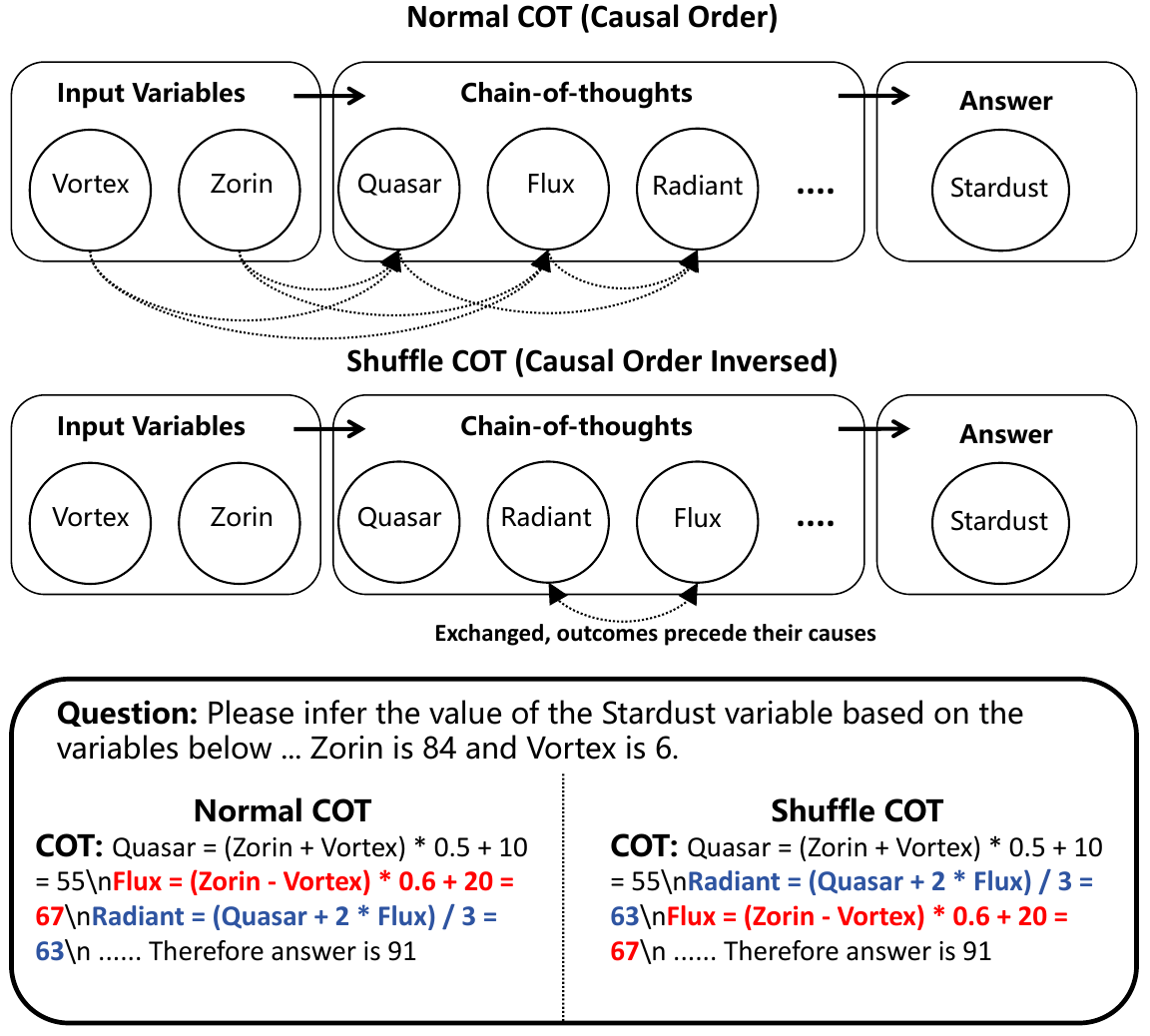}
    \caption{Normal COT follows the causal topological order of the data-generating process to construct reasoning steps, whereas the Shuffle setting simulates cases where COT exhibits causal misordering.}
    \label{fig:cot-shuffle}
\end{figure}

To verify the impact of AR and DLM attention limitations, we propose the COT-OrderPerturb synthetic dataset, thereby quantifying the impact of misalignment between the attention mechanism’s modeling priors on natural language and the causal priors underlying natural language. We first generate CoT simulation data based on a given prior causal graph, as shown in Figure \ref{fig:cot-shuffle}, including both CoTs that follow the standard causal order and CoTs with permuted concept sequences, where outcomes may precede their causes. To systematically explore different types of perturbations, we apply the following shuffling strategies: DFS, local reverse (LR), output first (OF), reverse (RE), and three random shuffles R$_1$, R$_2$, and R$_3$ (details in the Appendix \ref{APP:shuffle}), along with a control condition named No COT in which answers are generated directly without CoT reasoning. Results are summarized in Table \ref{tab:COT-Perturb}.

\begin{table*}[h!]
\centering
\resizebox{1\textwidth}{!}{
\begin{tabular}{l|c|cccccccc|c}
\toprule
\textbf{\multirow{2}{*}{Model}} &
\textbf{\multirow{2}{*}{Normal COT}} &
\multicolumn{8}{c|}{\textbf{Shuffle COT}} &
\textbf{\multirow{2}{*}{No COT}} \\
\cline{3-10}
 &  & \textbf{DFS} & \textbf{LR} & \textbf{OF} & \textbf{RE} & \textbf{R$_1$} & \textbf{R$_2$} & \textbf{R$_3$} &\textbf{avg$\pm$std} &  \\
\midrule
\rowcolor{gray!20} \textit{AR} & & & & & & & & & & \\
\textbf{Llama-3.2-1B} & 22.40\% & 20.60\% & 25.80\% & 31.40\% & 22.60\% & 25.00\% & 24.00\% & 24.20\% &24.80\%$\pm$3.37\%& 15.60\% \\
\textbf{Qwen3-8B} & \textbf{60.60\%} & 2.41\% & 44.20\% & 0.20\% & 0.20\% & 23.00\% & 32.40\% & 33.20\% &19.37\%$\pm$18.32\%& 36.40\% \\
\textbf{Llama-3.1-8B} & 47.60\% & 18.20\% & \textbf{44.00\%} & 14.00\% & 21.40\% & 4.00\% & 32.60\% & 29.80\% &23.43\%$\pm$13.20\%& 44.60\% \\

\midrule
\rowcolor{gray!20} \textit{DLM} & & & & & & & & &  &\\
\textbf{LLaDA-8B-Instruct (SFT)} & 38.60\% & \textbf{38.20\%} & 36.60\% & \textbf{42.40\%} & \textbf{41.80\%} & \textbf{33.80\%} & \textbf{35.00\%} & \textbf{40.60\%} &\textbf{38.34\%$\pm$3.38\%}& \textbf{57.60\%} \\
\bottomrule
\end{tabular}
}
\caption{Performance of AR and DLM under different setting in COT-OrderPerturb dataset.}\label{tab:COT-Perturb}
\end{table*}

\textbf{Structural Bias in AR Models.}
We observe that AR models exhibit declining performance consistency when data is perturbed such that outcomes precede their causes. However, linguistic flexibility is not bound to a strict left-to-right, token-by-token causal order, e.g., "The ground is slippery today because it rained" or "Lung cancer is caused by smoking". The outcome may precede the cause. The misalignment between AR priors and underlying causal priors of natural language introduces structural risks that cannot be resolved by simply scaling the training data.

\textbf{Robustness from Order-Independence in DLMs.} DLM, trained with fully connected attention and order-independent masking, demonstrates greater robustness. In the shuffled CoT setting, DLM achieves both a better mean and standard deviation of accuracy.

\textbf{Performance Bottlenecks of DLMs.}
Interestingly, DLMs perform notably worse than AR models under the Normal CoT setting. While AR models consistently benefit from CoT supervision, DLMs achieve substantially higher performance in the No-CoT condition than when CoT is included. This discrepancy arises because, by discarding causal order, DLM training effectively becomes a form of multi-objective learning across all reasoning steps. With limited data, this hinders the acquisition of deeper reasoning capabilities. Moreover, the longer CoT sequences further exacerbate inefficiency: DLMs typically require nearly 80 epochs to converge, whereas AR models converge within only 4 epochs.
\begin{figure}[h]
    \centering
    \includegraphics[width=0.8\linewidth]{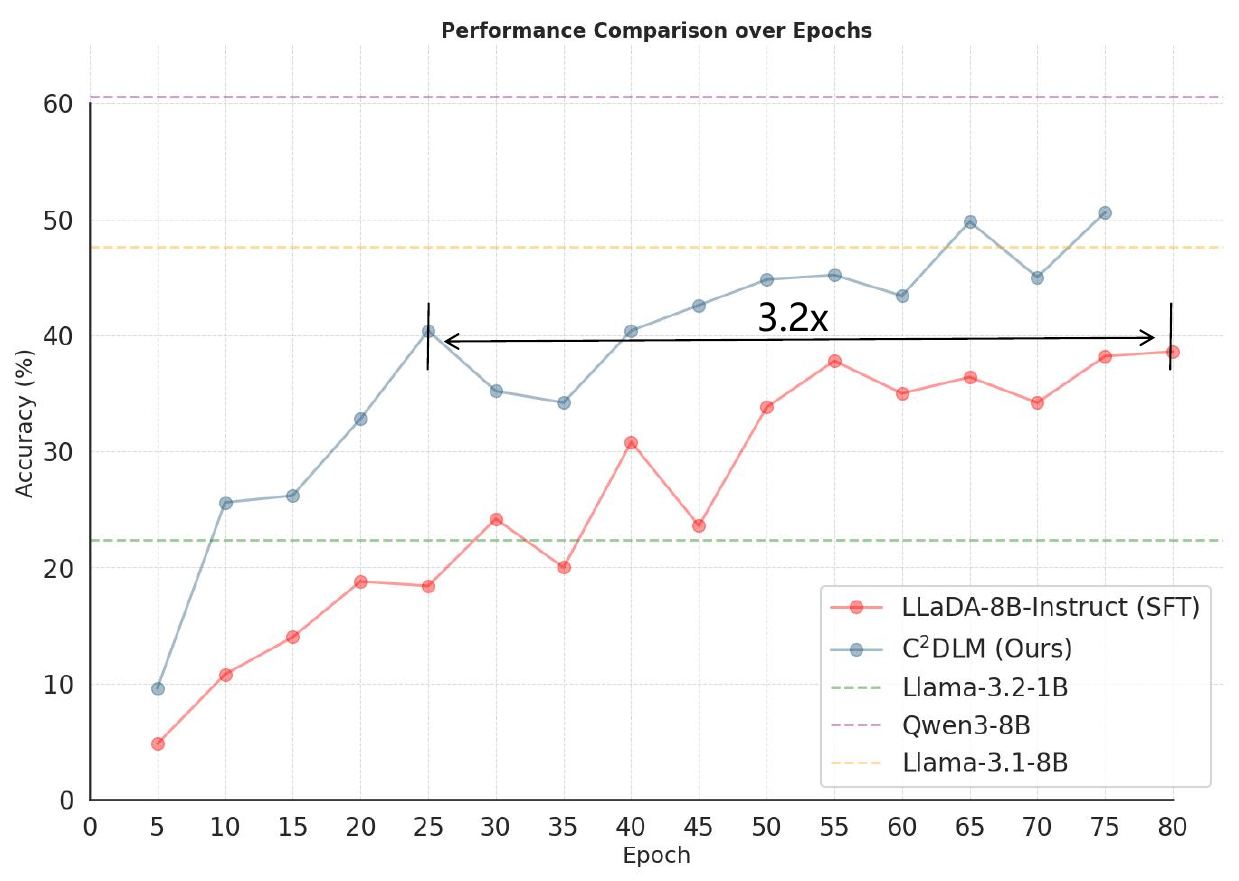}
    \caption{Accuracy curve as training progresses in the COT-OrderPerturb task.}
    \label{fig:zhexiantu}
\end{figure}

\begin{table}[h!]
\centering
\resizebox{0.35\textwidth}{!}{
\begin{tabular}{l|c}
\toprule
\textbf{Model} & \textbf{Normal COT} \\
\midrule
\textbf{LLaDA-8B-Instruct (SFT)} & 38.60\% \\
\textbf{C$^2$DLM (ours)} & \textbf{50.60\%}\\
\cellcolor{pink}$\Delta$ & \cellcolor{pink}+12.00\%\\
\bottomrule
\end{tabular}
}
\caption{Performance comparison under the Normal COT setting. Notation $\Delta$ denotes the performance gain relative to direct SFT on LLaDA-8B-Instruct.}\label{tab:normalcot}
\end{table}
\textbf{Performance Gain Using C$^2$DLM.} C$^2$DLM explicitly incorporates step-wise conceptual causal relationships, guiding the model to learn the data generation process via a V-aware Re-attention mechanism, and suppresses attention on element interactions that violate the causal structure. As shown in Table \ref{tab:normalcot}, aligning with causal priors leads to a significant performance improvement of 12\%, surpassing that of Llama-3.1-8B. Moreover, as illustrated in Figure \ref{fig:zhexiantu}, the training efficiency of C$^2$DLM is 3.2 times that of DLM.

\subsection{Downstream Task Experiments with Explicit Causal Structures}

\begin{table}[h!]
\centering

\resizebox{0.5\textwidth}{!}{
\begin{tabular}{l|ccc}
\toprule
\multirow{2}{*}{\textbf{Setting}}
 & \multicolumn{3}{c}{\textbf{Sudoku}} \\
\cmidrule(lr){2-4}
 & \textbf{n=200} & \textbf{n=500} & \textbf{n=5000} \\
\midrule
\rowcolor{gray!20} \textit{AR} & & &\\
\textbf{Llama-3.2-1B}   & 3.00\% & 22.40\% & 80.60\% \\
\textbf{Qwen3-8B}   & 67.40\% & 76.40\% & 83.00\% \\
\textbf{Llama-3.1-8B}   & 8.60\% & 29.20\% & 80.80\% \\

\midrule
\rowcolor{gray!20} \textit{DLM} & & &\\
\textbf{LLaDA-8B-Instruct (SFT)}   & 77.05\% & 90.23\% & 92.14\% \\
\textbf{C$^2$DLM (ours)}  & \textbf{87.89\%} & \textbf{91.21\%} & \textbf{92.97\%} \\
\cellcolor{pink}$\Delta$ & \cellcolor{pink}{+10.84\%} & \cellcolor{pink}{+0.98\%} & \cellcolor{pink}{+0.83\%} \\

\bottomrule
\end{tabular}
}
\caption{Performance on Sudoku task. And n represents the size of the training data. Notation $\Delta$ denotes the performance gain relative to direct SFT on LLaDA-8B-Instruct.}\label{tab:sudoku}
\end{table}

\begin{table*}[h!]
\centering
\resizebox{\textwidth}{!}{
\begin{tabular}{l|cc|cc|cc|cc}
\toprule
\multirow{2}{*}{\textbf{Setting}}
 & \multicolumn{2}{c}{\textbf{STG\_S}}
 & \multicolumn{2}{c}{\textbf{STG\_M}}
 & \multicolumn{2}{c}{\textbf{STG\_L}}
 & \multicolumn{2}{c}{\textbf{STG\_H}}\\
\cmidrule(lr){2-3}\cmidrule(lr){4-5}\cmidrule(lr){6-7}\cmidrule(lr){8-9}
 & \textbf{IID} & \textbf{OOD}
 & \textbf{IID} & \textbf{OOD}
 & \textbf{IID} & \textbf{OOD}
 & \textbf{IID} & \textbf{OOD} \\
\midrule
\rowcolor{gray!20} \textit{AR} & & & & & & & &\\
\textbf{Llama-3.2-1B}   & 83.50\% & 67.25\%
        & \textbf{95.75\%} & 61.50\%
        & \textbf{97.25\%} & 87.00\%
        & 37.40\% & 27.90\% \\
\textbf{Qwen2.5-1.5B\dag}   & 81.50\% & 78.50\%
        & 93.25\% & 82.00\%
        & 95.75\% & 82.00\%
        & 51.50\% & \textbf{53.40\%} \\
\textbf{Llama-3.1-8B\dag}   & \textbf{90.50\%} & 86.25\%
        & 93.25\% & 64.50\%
        & 96.00\% & 88.25\%
        & \textbf{57.80\%} & 49.60\% \\
\midrule
\rowcolor{gray!20} \textit{DLM} & & & & & & & &\\
\textbf{LLaDA-8B-Instruct (SFT)}   & 86.50\% & 81.25\%
        & 85.75\% & 83.00\%
        & 88.25\% & 83.25\%
        & 24.80\% & 25.60\% \\
\textbf{C$^2$DLM (ours)}  & 88.00\% & \textbf{88.50\%}
               & 93.00\% & \textbf{95.00\%}
               & 93.50\% & \textbf{92.25\%}
               & 35.20\% & 32.40\% \\
\cellcolor{pink}$\Delta$ & \cellcolor{pink}{+1.50\%} & \cellcolor{pink}{+7.25\%}
         & \cellcolor{pink}{+7.25\%} & \cellcolor{pink}{+12.00\%}
         & \cellcolor{pink}{+5.25\%} & \cellcolor{pink}{+9.00\%}
         & \cellcolor{pink}{+10.40\%} & \cellcolor{pink}{+6.80\%} \\
\bottomrule
\end{tabular}
}
\caption{Performance on STG task. Notation "\dag" means results from  \cite{han2025cat}. Notation $\Delta$ denotes the performance gain relative to direct SFT on LLaDA-8B-Instruct.}\label{tab:stg}
\end{table*}

Some downstream tasks contain explicit causal structures as priors in their data generation processes. We selected Sudoku and STG as representative benchmarks to evaluate C$^2$DLM.

\subsubsection{Sudoku dataset}
For the Sudoku task, we focus on a $4 \times 4$ grid, which aligns the setting the same as  \cite{zhao2025d1}. In Sudoku, each number in Sudoku is determined by the values in its row, column, and corresponding subgrid. Experimental results in Figure \ref{tab:sudoku} indicate that AR approaches are constrained by the unidirectional flow of information, causing a misalignment between attention's prior and the data generation process, which leads to low performance. By contrast, C$^2$DLM effectively leverages the causal priors, which allows the model to better fit the data and avoid learning spurious and unrelated correlations. This advantage is particularly obvious in small-data scenarios ($n=200$), where LLaDA-8B-Instruct lacks clear supervisory guidance and performs worse than C$^2$DLM.

\subsubsection{STG dataset}
\begin{figure}
    \centering
    \includegraphics[width=1\linewidth]{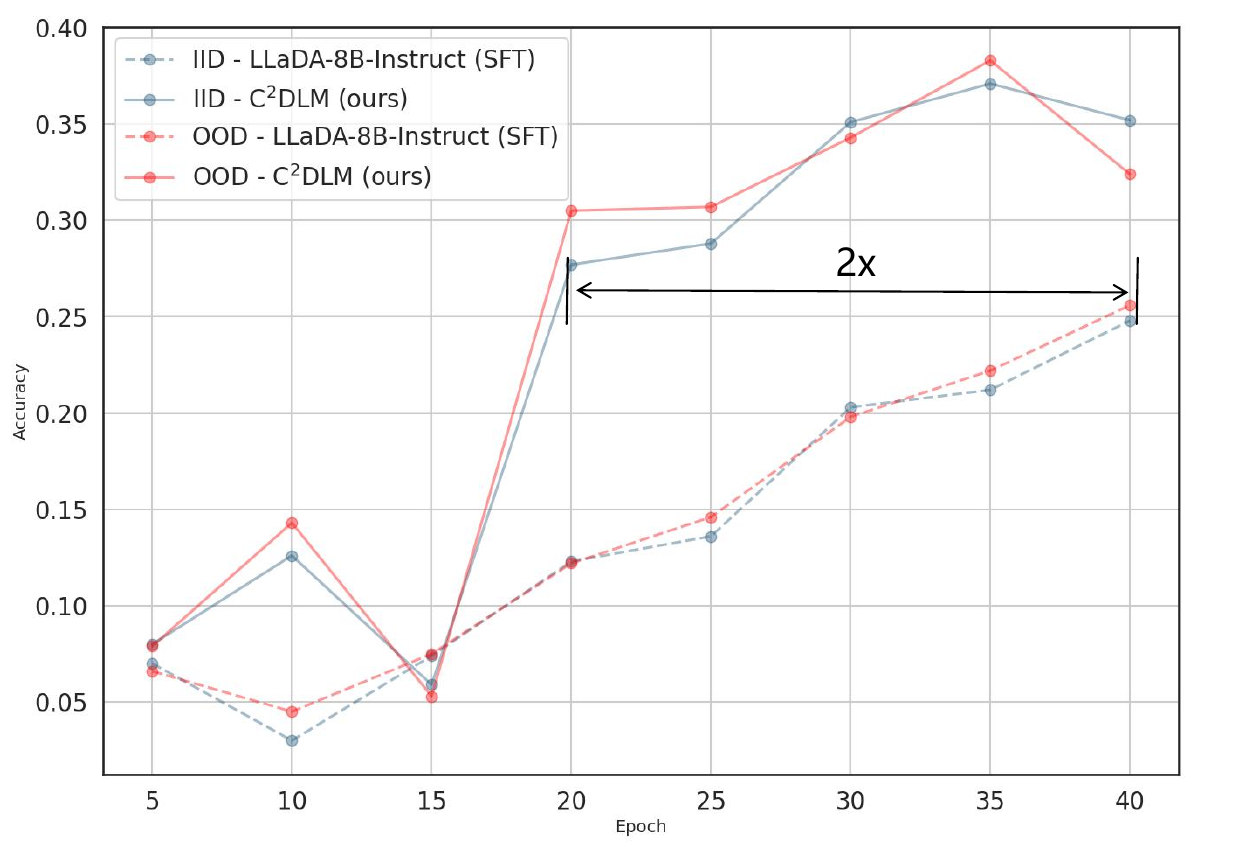}
    \caption{Performance change curve during different
epochs of training on the STG\_H dataset.}
    \label{fig:stg_h}
\end{figure}

Similarly, the STG dataset is also generated from explicit causal graphs and provides both IID and OOD testing scenarios, which facilitate a more systematic evaluation of robustness in the presence of spurious correlations. As shown in \ref{tab:stg}, C$^2$DLM significantly outperforms direct SFT across different STG subsets, with an average improvement of 7.43\% over IID and OOD settings. Compared to AR, introducing the causal prior via C$^2$DLM markedly narrows their gap and even surpasses the best AR baselines in the OOD setting of STG\_S, STG\_M, and STG\_L.

We further examined the training efficiency of C$^2$DLM in STG\_H. As shown in Figure \ref{fig:stg_h}. The training efficiency of C$^2$DLM is 2 times that of DLM.

\begin{figure}[h!]
    \centering
    \includegraphics[width=1\linewidth]{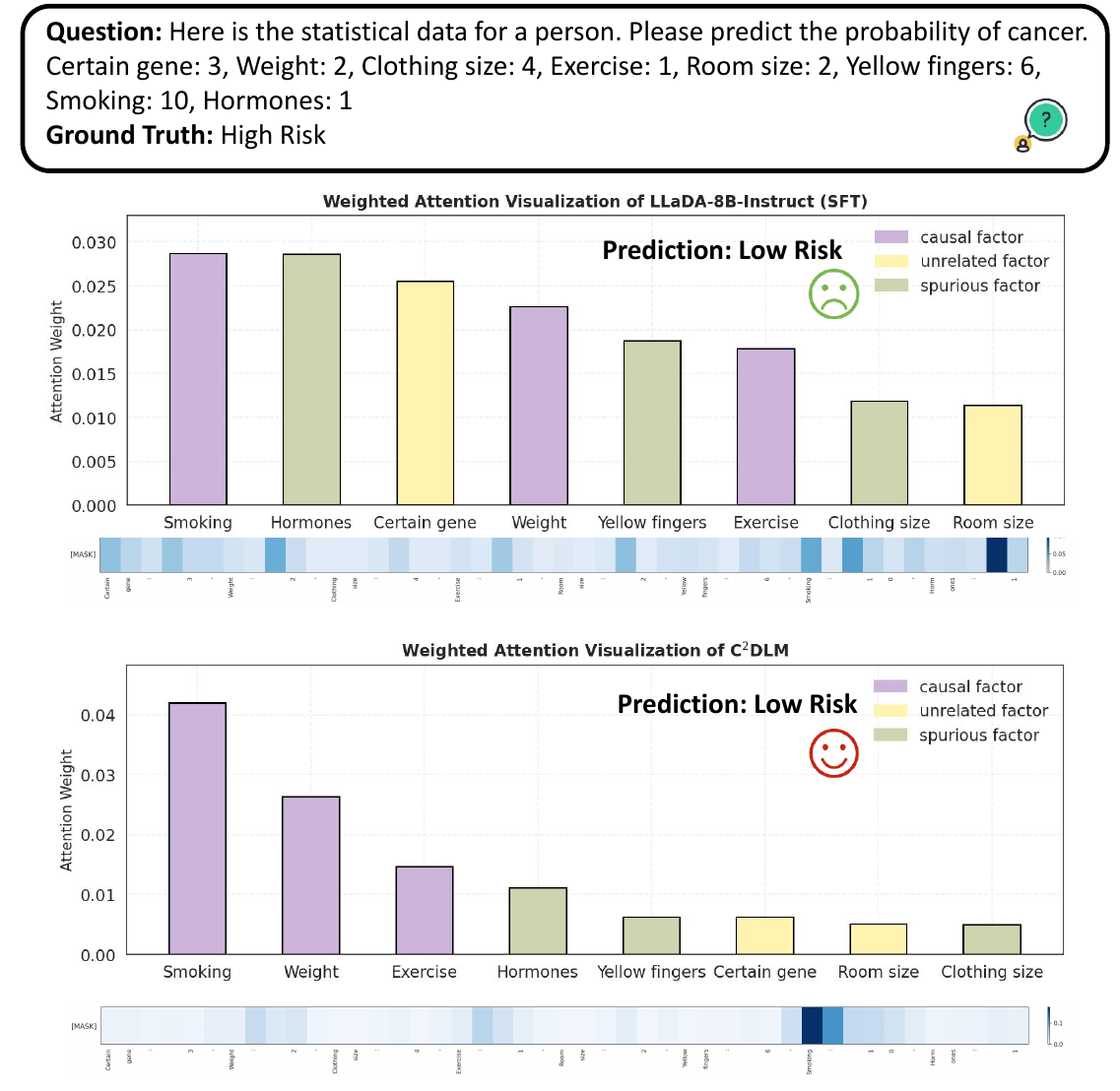}
    \caption{Attention visualization and weight distribution bar charts, where the x-axis represents different attributes in the STG task. Purple indicates causal factors, green denotes spurious correlations, and yellow represents unrelated factors.}
    \label{fig:keshihua}
\end{figure}

To better interpret the impact of C$^2$DLM on the attention mechanism, we conducted attention visualizations weighted by the value matrix. As shown in Figure \ref{fig:keshihua}, C$^2$DLM effectively learns causal relationships, whereas direct fine-tuning of LLaDA fails to distinguish spurious correlations and irrelevant factors from causal factors. By mechanistically guiding the model to fit the data generation process, C$^2$DLM  yields more robust predictions and enhances model reliability. Further analysis is provided in Appendix \ref{APP:keshihua}.

\subsection{Evaluate on Broader  Math and Reasoning Downstream Tasks}
\begin{table*}[h!]
\centering

\resizebox{\textwidth}{!}{
\begin{tabular}{l|cccccc|c}
\toprule
\textbf{Setting} & \textbf{GSM8K} & \textbf{MATH500} & \textbf{GPQA} & \textbf{MMLU\_STEM} & \textbf{ARC\_C} & \textbf{SAT} & \textbf{Avg} \\
\midrule
\textbf{LLaDA-8B-Instruct} & 80.36\% & 36.60\% & 28.79\% & 58.74\% & 85.75\% & 71.36\% & 60.27\% \\
\textbf{LLaDA-8B-Instruct (SFT)}               & 80.74\% & 36.20\% & 29.24\% & 59.21\% & 85.67\% & 75.00\% & 61.01\% \\
\textbf{C$^2$DLM (ours)}         & \textbf{81.96\%} & \textbf{37.20\%} & \textbf{29.46\%} & \textbf{60.42\%} & \textbf{86.26\%} & \textbf{78.64\%} & \textbf{62.32\%} \\
\midrule
\cellcolor{pink}$\Delta$          &\cellcolor{pink}+1.22\% & \cellcolor{pink}+1.00\% & \cellcolor{pink}+0.22\% & \cellcolor{pink}+1.21\% & \cellcolor{pink}+0.59\% & \cellcolor{pink}+3.64\% & \cellcolor{pink}+1.31\% \\
\bottomrule
\end{tabular}
}
\caption{Performance on diverse math and reasoning downstream datasets. Notation $\Delta$ denotes the performance gain relative to direct SFT on LLaDA-8B-Instruct.}\label{tab:downstream}
\vspace{-1em}
\end{table*}

For broader downstream tasks without explicit causal structure, we leverage the automated workflow introduced in Section \ref{sec:Concept-level}. Specifically, we adopt the latest GLM-4.5\cite{zeng2025glm}, which provides a balance between scalability and cost-effectiveness while addressing the challenges of extracting causal relationships from long CoT sequences. Aligning with previous work\cite{zhao2025d1,tang2025wd1}, we start from the s1k dataset and construct a training dataset containing 686 instances annotated by the GLM-4.5.

We conducted a manual random sampling of 50 instances to evaluate the causal graphs generated by the teacher model. Two instances (4\% of the sampled data) failed to produce causal graphs due to decoding errors. Among the successfully decoded instances, the accuracy was 93.42\% ± 1.41\%. Detailed experimental procedures are provided in Appendix \ref{APP:Human}.

Both SFT and C$^2$DLM are trained on this dataset, with the only difference being that C$^2$DLM incorporates the causal prior loss. The resulting models are then evaluated on six downstream tasks. As shown in Table \ref{tab:downstream}, C$^2$DLM achieves consistent improvements across six test datasets, with an average gain of 1.31\%. Notably, these gains are obtained using only 686 causally annotated examples. As performance improvements from scaling next-token prediction alone are approaching a bottleneck, our pipeline highlights the potential of leveraging two-dimensional supervision signals based on token interactions as a promising direction for future scaling, with cost as low as 0.46\$ per million tokens (see Appendix \ref{APP:cost} for details).

\subsection{Ablation Study}

\begin{table}[h!]
\centering

\resizebox{0.5\textwidth}{!}{
\begin{tabular}{l|ccc|c}
\toprule
\textbf{Setting} & \textbf{GSM8K} & \textbf{MATH500} & \textbf{SAT} & \textbf{Avg} \\
\midrule
\textbf{$\alpha=2$} & 81.43\% & \textbf{37.60\%} & 77.73\% & 65.58\% \\
\cellcolor{gray!30}\textbf{$\alpha=3$} & \cellcolor{gray!30}\textbf{81.96\%} &\cellcolor{gray!30}
37.20\%&\cellcolor{gray!30} \textbf{78.64\%} & \cellcolor{gray!30} \textbf{65.93\%}\\

\quad \textbf{w/o $\mathcal{S}_\gamma$} & 81.65\% & 33.20\% & 74.09\% & 62.98\% \\
\quad \textbf{w/o} V-aware & 81.35\% & 34.00\% & 68.64\% & 61.33\% \\

\textbf{$\alpha=4$} &81.65\% & 34.00\% & 76.82\% & 64.16\% \\

\textbf{$\alpha=5$} & 81.50\% & 36.80\% & 78.18\% & 65.49\% \\
\bottomrule
\end{tabular}
}
\caption{Ablation study under different $\alpha$ and $\gamma$ scheduler. Gray line ($\alpha=3$) is the default setting. And w/o $\mathcal{S}_\gamma$ means don't use $\gamma$ scheduler.}\label{tab:ablation}
\vspace{-1em}
\end{table}

To analyze the effects of different components and hyperparameters, we conducted ablation studies on the parameters $\alpha$ and $\gamma$ scheduler. $\alpha$ represents the degree of emphasis on causal relationships, and an appropriate level of emphasis contributes to improved model performance. Additionally, we evaluated an ablation of the V-aware strategy, in which the model performance was assessed without the weighting provided by the Value matrix (denoted as w/o V-aware). As shown in Table \ref{tab:ablation}, within a certain range, model performance first improves and then declines as $\alpha$ increases. We further examined the impact of the $\gamma$ scheduler. Without $\gamma$ scheduler, the performance of C$^2$DLM declines on all datasets. When the V-aware strategy is not employed, and causal knowledge is directly injected, performance decreases. This is due to the direct manipulation of attention scores. Ignoring the influence of the Value matrix introduces substantial instability during training, which in turn leads to performance degradation. These results demonstrate the effectiveness of the V-aware strategy.

\section{Conclusion}
To address the limitations of language models in reasoning, we propose C$^2$DLM, a new paradigm distinct from AR and DLM. C$^2$DLM leverages automated pipelines to extract causal meta-knowledge and employs the V-aware Re-attention mechanism to align attention. We propose the COT-OrderPerturb task to quantify the influence of language modeling priors, and we validate the effectiveness of C$^2$DLM on Sudoku, STG, and six broad downstream tasks. C$^2$DLM improves both the model’s reasoning ability and training efficiency. Furthermore, we reveal the risk of misalignment between attention mechanisms and the causal priors underlying natural language, which shows the potential of combining causality into language models.
\section*{Limitations}
Our experiments focus on the LLaDA-8B-Instruct model, but due to constraints in training resources and the base model, C$^2$DLM’s performance still lags behind the SOTA AR models. Furthermore, larger-scale DLMs remain underexplored, so the effectiveness of C$^2$DLM on such models is still unknown. Limited by computational resources, we are unable to inject causal knowledge at scale during pretraining; the pretraining stage remains largely unexamined, and our current work focuses solely on the SFT phase. Causal knowledge in the real world is highly complex, and thus extracting causal structures and benefiting from them in more intricate causal graphs or ultra-long CoT remains a significant challenge.

\section*{Ethical considerations}

A potential risk of this work lies in the possibility that the proposed method could be misused to inject illegal or unethical information into models. Therefore, we strongly urge users to ensure that all training datasets comply with relevant laws and ethical guidelines when applying this approach.

\bibliography{custom}

\appendix

\section{Details of Prompt}\label{APP:Prompt}

This appendix provides the prompt templates used for concept-level causal meta-knowledge extraction with a Teacher LLM. By decomposing questions and extracting concept-level causal relations, we derive reasoning graphs, which serve as a surrogate for the underlying causal priors of natural language.

\newtcolorbox[auto counter, number within=section]{mybox}[2][]{
  colframe=green!50!black,
  colback=white,
  coltitle=black,
  fonttitle=\bfseries,
  title={#2},
  sharp corners,
  boxrule=0.4pt,
  breakable,
  #1
}
\begin{mybox}[title={Teacher LLM's System Prompt}]

\#\#Command:
You are a causality expert. Your task is to identify causal relationships in a given problem and its solution, in order to generate a two-dimensional supervisory signal between tokens. Specifically, you need to extract concepts in the chain of thought. These concepts should:

1.Reflect the essential reasoning process.

2.Carry richer information content.

3.Have a stronger impact on predicting the final answer.

4.Be unique within the text, to avoid ambiguity during matching.

5.The extracted content must come strictly from the original text provided.

The extraction format should be a dictionary: \{ "Effect1": [Cause11, Cause12, ...], "Effect2": [Cause21, Cause22, ...], ... \}.The dictionary key concept represents the effect, while the value list represents the causes. Therefore, in each segment of text, you need to carefully determine which concept is the cause and which is the effect — that is, without the cause, the effect cannot be naturally obtained. It is best if the concepts can be linked end-to-end through causal relationships, forming a complete reasoning map.
\end{mybox}

\begin{mybox}[title={Teacher LLM's In-context Demo}]

Question: Given a rational number, write it as a fraction in lowest terms and calculate the product of the resulting numerator and denominator. For how many rational numbers between 0 and 1 will $20!$ be the resulting product?

Answer: Given a rational number between 0 and 1, we need to write it as a fraction in lowest terms \(\frac{a}{b}\) and calculate the product of the resulting numerator and denominator. We need to determine how many such rational numbers exist where the product \(a \times b = 20!\). \\

1. \textbf{Prime Factorization of 20!:} \\
- The prime factors of 20! are 2, 3, 5, 7, 11, 13, 17, and 19. These are 8 distinct primes. \\

2. \textbf{Coprime Pairs:} \\
- For \(a \times b = 20!\) and \(\gcd(a, b) = 1\), each prime factor of 20! must be assigned entirely to either \(a\) or \(b\). This results in \(2^8 = 256\) coprime pairs \((a, b)\). \\

3. \textbf{Counting Pairs \((a < b)\):} \\
- Since \(a\) and \(b\) are coprime and \(a \times b = 20!\), each pair \((a, b)\) and \((b, a)\) are distinct. To find pairs where \(a < b\), we divide the total number of pairs by 2, excluding the case \(a = b\) (which is impossible since 20! is not a square). \\

4. \textbf{Result:} \\
- The number of coprime pairs \((a, b)\) with \(a < b\) is \(\tfrac{256}{2} = 128\). \\

Thus, the number of rational numbers between 0 and 1 for which the product of the numerator and denominator in lowest terms is \(20!\) is \(\boxed{128}\). \\

\#\# Demo Output 1: \\
\{ \\
``2, 3, 5, 7, 11, 13, 17, and 19": ["The prime factors of 20!"], \\
"This results in \(\,2^8 = 256\) coprime pairs \((a, b)\)": ["2, 3, 5, 7, 11, 13, 17, and 19", "each prime factor of 20! must be assigned entirely to either \((a)\) or \((b)\)"], \\
"each pair \((a, b)\) and \((b, a)\) are distinct": ["Since \((a)\) and \((b)\) are coprime and \(a \times b = 20!\)"], \\
"The number of coprime pairs \((a, b)\) with \(a < b\) is \(\tfrac{256}{2} = 128\)" : ["This results in \(\,2^8 = 256\) coprime pairs \((a, b)\)", "we divide the total number of pairs by 2, excluding the case \((a = b)\)"] \\
\} \\

\#\# Demo Input 2: \\
Question: Square $AIME$ has sides of length $10$ units.  Isosceles triangle $GEM$ has base $EM$, and the area common to triangle $GEM$ and square $AIME$ is $80$ square units. Find the length of the altitude to $EM$ in \(\triangle GEM\).

Answer: Square \((AIME)\) has sides of length 10 units. Isosceles triangle \((GEM)\) has base \((EM)\), and the area common to triangle \((GEM)\) and square \((AIME)\) is 80 square units. We need to find the length of the altitude to \((EM)\) in \(\triangle GEM\). \\

1. \textbf{Vertices of the Square:} \\
- Place the square on a coordinate system with vertices \((A(0, 0))\), \((I(0, 10))\), \((M(10, 10))\), and \((E(10, 0))\). \\

2. \textbf{Isosceles Triangle \((GEM)\):} \\
- Base \((EM)\) is vertical from \((E(10, 0))\) to \((M(10, 10))\). \\
- The apex \(G\) of the triangle is to the left of \((EM)\) (outside the square for larger altitudes). \\

3. \textbf{Coordinates of \(G\):} \\
- Let the altitude from \(G\) to \((EM)\) be \(h\). The coordinates of \(G\) are \((10 - h, 5)\) because the triangle is isosceles with \((GE = GM)\). \\

4. \textbf{Equations of Lines:} \\
- Line \((GE)\) has the equation \(y = -\tfrac{5}{h}x + \tfrac{50}{h}\). \\
- Line \((GM)\) has the equation \(y = \tfrac{5}{h}x + 10 - \tfrac{50}{h}\). \\

5. \textbf{Intersection with the Square:} \\
- The lines \((GE)\) and \((GM)\) intersect the left edge of the square (x=0) at points \((0, \tfrac{50}{h})\) and \((0, 10 - \tfrac{50}{h})\). \\

6. \textbf{Area Calculation:} \\
- The overlap area is: \\
\begin{align*}
\text{Area}
&= \int_{0}^{10}
\Biggl( \left( \tfrac{5}{h}x + 10 - \tfrac{50}{h} \right) \\
&\quad - \left( -\tfrac{5}{h}x + \tfrac{50}{h} \right) \Biggr) dx
\end{align*}
- Simplifying: \(\text{Area} = \int_{0}^{10} \left( \tfrac{10}{h}x + 10 - \tfrac{100}{h} \right) dx = 100 - \tfrac{500}{h}\). \\
- Setting \(\text{Area} = 80\): \(100 - \tfrac{500}{h} = 80 \implies h = 25\). \\

Thus, the length of the altitude to \((EM)\) in \(\triangle GEM\) is \(\boxed{25}\). \\

\#\# Demo Output 2: \\
\{ \\
"The coordinates of \((G)\) are \((10 - h, 5)\)": ["the triangle is isosceles with \((GE = GM)\)", "Let the altitude from \((G)\) to \((EM)\) be \((h)\)", "Place the square on a coordinate system with vertices \((A(0, 0))\), \((I(0, 10))\), \((M(10, 10))\), and \((E(10, 0))\)"], \\
"The lines \((GE)\) and \((GM)\) intersect the left edge of the square (x=0) at points \((0, \tfrac{50}{h})\) and \((0, 10 - \tfrac{50}{h})\)": ["Line \((GE)\) has the equation \(y = -\tfrac{5}{h}x + \tfrac{50}{h}\)", "Line \((GM)\) has the equation \(y = \tfrac{5}{h}x + 10 - \tfrac{50}{h}\)"], \\
"the length of the altitude to \((EM)\) in \(\triangle GEM\) is \(\boxed{25}\)." : ["Setting the area equal to 80: \(100 - \tfrac{500}{h} = 80 \implies h = 25\)"] \\
\}
\end{mybox}

\label{sec:appendix}

\section{Details of Hyperparameters}\label{APP:Hyper}

Since different tasks vary in sequence length and convergence speed, we assign task-specific training epochs and generation lengths at evaluation. The detailed configurations are as follows:

For the COT-OrderPerturb task, we train for 10 epochs with a generation length of 512. To fully examine the impact of the DLM generation paradigm without any AR strategy, we additionally set the block length to 512.

For the Sudoku task, we train for 10 epochs with a generation length of 256 at evaluation.

For the STG task, the STG\_E subset is a binary classification problem, and we therefore train for 10 epochs. The more challenging STG\_H subset is trained for 40 epochs. Since these tasks do not involve chain-of-thought reasoning, the generation length is fixed at 8.

For the six downstream tasks, we follow the setup of  \cite{zhao2025d1,tang2025wd1} and train on the s1k-1.1 subset\footnote{https://huggingface.co/datasets/simplescaling/s1K-1.1}, with the training context length set to 1600. At evaluation, the generation length is set to 512 for GSM8K, MATH500, and SAT, while all other choice tasks use a generation length of 32.

For the autoregressive model baselines, we uniformly adopt a LoRA learning rate of $2 \times e^{-4}$. Sudoku is trained for 8 epochs, while all other tasks are trained for 4 epochs.

All training is conducted on 2 $\times$ NVIDIA A100 40GB GPUs, with the random seed fixed at 42 across all experiments. For testing, GSM8K, MATH500, and SAT are run on 4 $\times$ A100 40GB GPUs, while all other tasks are evaluated on 2 $\times$ A100 40GB GPUs.

Detailed implementation examples can be found in our code repository.

\section{Generation Details of COT-OrderPerturb Dataset}\label{APP:shuffle}

The goal of this process is to generate chain-of-thought reasoning trajectories with controlled order perturbations while preserving the underlying causal structure.

\subsection{Graph Construction and Chain-of-Thought Generation}
We define a directed acyclic graph (DAG) template in which nodes represent abstract variables (e.g., Quasar, Flux, Radiant, Nova), and edges encode functional dependencies. Each non-source variable is associated with a deterministic transformation rule, typically a linear or nonlinear combination of its parent variables. Two source nodes (Zorin and Vortex) are sampled uniformly from the integer range [0,100], while all other variables are computed sequentially via topological ordering. The target variable Stardust is uniquely determined by this process, ensuring consistency across samples.

Given the DAG and computed values, we construct step-wise reasoning traces. Each reasoning step includes: the functional rule applied (e.g., Quasar = (Zorin + Vortex) * 0.5 + 10), the input variables, and the evaluated output.

\begin{figure}[h!]
    \centering
    \includegraphics[width=1\linewidth]{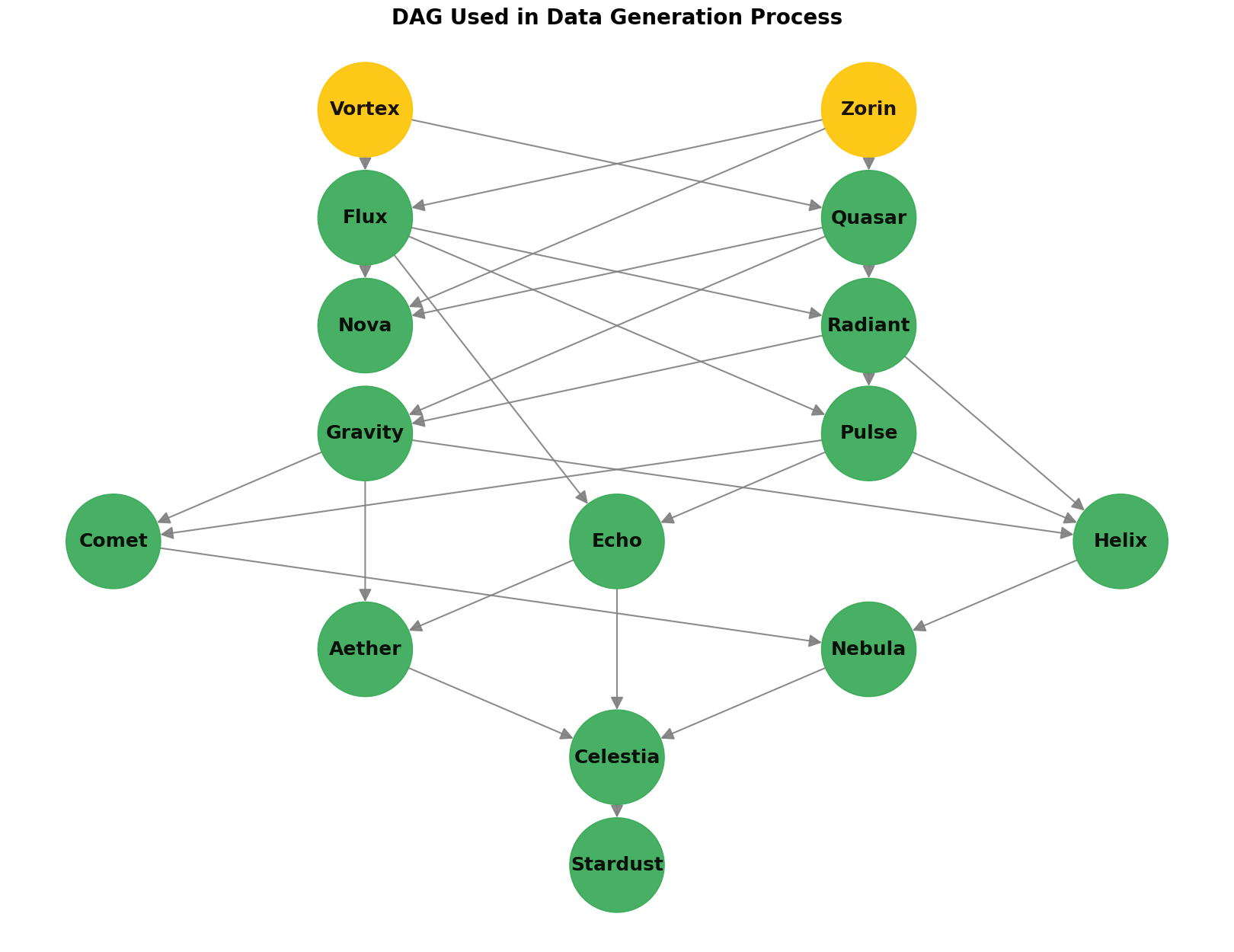}
    \caption{DAG used in data generation process. Yellow nodes are input variables. The Stardust variable is the final answer.}
    \label{fig:dag}
\end{figure}

When arranged in strict topological order, these steps form a reasoning trajectory that faithfully reflects the data-generating process. Finally, the data generation process is shown in Figure \ref{fig:dag}. One example is as following:

\begin{mybox}[title={COT-OrderPerturb Example}]

\textbf{Question}: Please infer the value of the \texttt{Stardust} variable based on the variables below.
The input variables are \texttt{Zorin} (value: 80) and \texttt{Vortex} (value: 79). \\[6pt]

\textbf{COT}: \\
\quad Quasar = (Zorin + Vortex) * 0.5 + 10 = 90 \\
\quad Flux = (Zorin - Vortex) * 0.6 + 20 = 21 \\
\quad Radiant = (Quasar + 2 * Flux) / 3 = 44 \\
\quad Nova = (Quasar - Flux + Zorin) / 3 + 5 = 55 \\
\quad Gravity = (Radiant * Quasar) / 120 + 8 = 41 \\
\quad Pulse = Radiant * 0.4 + Flux * 0.9 = 36 \\
\quad Helix = (Gravity + Pulse + Radiant) / 3 = 40 \\
\quad Echo = (Pulse - Flux) * 0.8 = 12 \\
\quad Comet = (Pulse + Gravity) * 0.6 + 2 = 48 \\
\quad Aether = (Echo + Gravity) * 0.5 = 26 \\
\quad Nebula = (Helix + Comet) / 2 + 3 = 47 \\
\quad Celestia = (Nebula + Aether + Echo) * 1.1 + 6 = 100 \\
\quad Stardust = int(Celestia * 0.7) = 70 \\[4pt]
Therefore, the final answer is 70. \\[6pt]

\textbf{Answer}: 70
\end{mybox}

\subsection{Order Perturbations}

To examine robustness to reasoning irregularities, we apply controlled perturbations to the canonical reasoning sequence. The perturbation modes are:

\begin{itemize}
    \item \textbf{reverse (RE)}: complete reversal of all steps.
    \item \textbf{local reverse (LR)}: pairwise reversal within local consecutive steps.
    \item \textbf{output first (OF)}: moving the final output computation to the beginning.
    \item \textbf{DFS}: depth-first–style ordering based on node depth.
    \item \textbf{random $\times 3$ (R1, R2, R3)}: three fixed random permutations of reasoning steps.
    \item \textbf{no cot}: omission of reasoning; only the final answer is given.
\end{itemize}

These perturbations preserve the correctness of the final answer (Stardust) while only shuffling the intermediate reasoning trajectory.

In total, 2000 unique base samples are generated, yielding multiple perturbed training subsets. We generate another 500 different samples as a unique test dataset. The pseudo-code of the data generation process is shown in Algorithm \ref{alg:cot_orderperturb}.
\begin{algorithm}[h]
\caption{COT-OrderPerturb Data Generation}
\label{alg:cot_orderperturb}
\begin{algorithmic}[1]
    \REQUIRE Number of base samples $N$, DAG template $\mathcal{G}$, Perturbation modes $\mathcal{M}$
    \ENSURE Test set $\mathcal{T}$, Training sets $\{\mathcal{D}_m\}_{m \in \mathcal{M}}$

    \STATE Initialize $\mathcal{T} \leftarrow \emptyset$, $\mathcal{D}_m \leftarrow \emptyset$ for all $m \in \mathcal{M}$
    \STATE Initialize seen signatures $\mathcal{S} \leftarrow \emptyset$
    \STATE Define perturbation modes: \texttt{RE}, \texttt{LR}, \texttt{OF},  \texttt{DFS}, \texttt{R$_1\dots_4$}, \texttt{no cot}

    \WHILE{$|\mathcal{T}| < N$}
        \STATE Sample source nodes $(Zorin, Vortex) \sim \text{Uniform}(0,100)$
        \STATE Evaluate all other nodes in $\mathcal{G}$ via topological order
        \STATE Compute final target variable $Stardust$
        \STATE Create signature $\sigma$ $ = $ $(Zorin, Vortex, Stardust)$
        \IF{$\sigma \in \mathcal{S}$}
            \STATE \textbf{continue}
        \ENDIF
        \STATE Add $\sigma$ to $\mathcal{S}$
        \STATE Construct canonical chain-of-thought steps $\pi$
        \STATE Store canonical sample $(Q, \pi, Stardust)$ into test set $\mathcal{T}$

        \FORALL{$m \in \mathcal{M}$}
            \STATE Apply perturbation $m$ to steps $\pi \to \pi_m$
            \STATE Store perturbed sample $(Q, \pi_m, Stardust)$ into $\mathcal{D}_m$
        \ENDFOR
    \ENDWHILE
    \STATE \textbf{return} $\mathcal{T}, \{\mathcal{D}_m\}_{m \in \mathcal{M}}$
\end{algorithmic}
\end{algorithm}

\section{Cost Analysis of C$^2$DLM}\label{APP:cost}

We also analyze the cost of using GLM-4.5 for causal annotation. Specifically, we randomly select 100 examples as a representative subset to estimate annotation costs and calculate both input and output token counts. For the input, the average token length is 865.2, and with an additional prompt overhead of 1981 tokens. For the output, the average length is 295.3 tokens.

Formally, the average cost for one token is computed as:

$$
\text{Cost} = \frac{1}{865.2}(
T_{\text{in}} \cdot P_{\text{in}} +
T_{\text{out}} \cdot P_{\text{out}}),
$$

where $T_{in}$ and $T_{out}$ denote the total input and output token counts. So $T_{in}=2846.2$, $T_{out}=295.3$. And the official pricing of GLM-4.5 is $P_{in}$=0.8 RMB/M tokens, $P_{out}$=2.0 RMB/M tokens.

Substituting the empirical statistics:

$$
\text{Cost} =
\frac{1}{865.2}\times(2846.2 \times 0.8 +
295.3 \times 2.0)
= 3.31.
$$

Converting to USD (1 RMB $\approx$ 0.14 USD)\footnote{The pricing unit of the GLM API is in CNY, approximately 1 CNY per million tokens, which is equivalent to about 0.14 USD based on the exchange rate as of October 2, 2025.}, the total annotation cost is about 0.46\$ per million tokens.
\section{Attention Visualization of C$^2$DLM}\label{APP:keshihua}
\begin{figure*}
    \centering
    \includegraphics[width=1\linewidth]{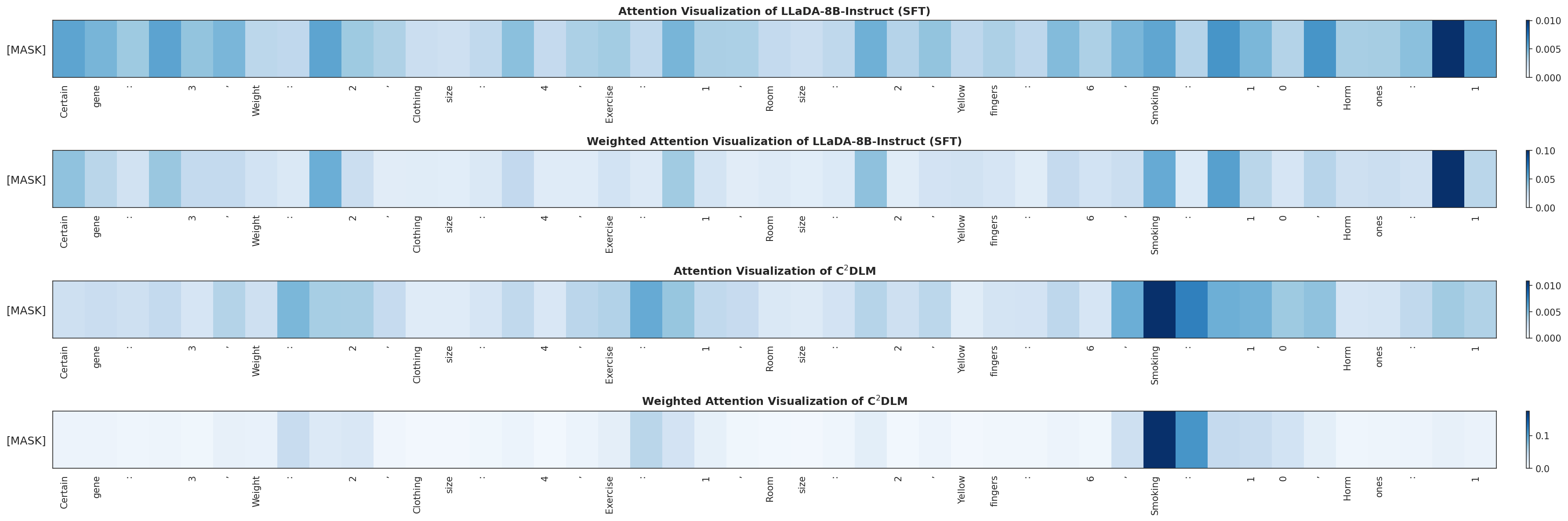}
    \caption{Mask is the position which will be decoded as high or low (directly determines the final answer). Visualization of the attention matrices, presented from top to bottom: direct visualization of LLaDA’s attention map; LLaDA’s attention map weighted by the Value norm; direct visualization of C$^2$DLM’s attention map; and C$^2$DLM’s attention map weighted by the Value norm.}
    \label{fig:app_keshihua1}
\end{figure*}

\begin{figure*}
    \centering
    \includegraphics[width=1\linewidth]{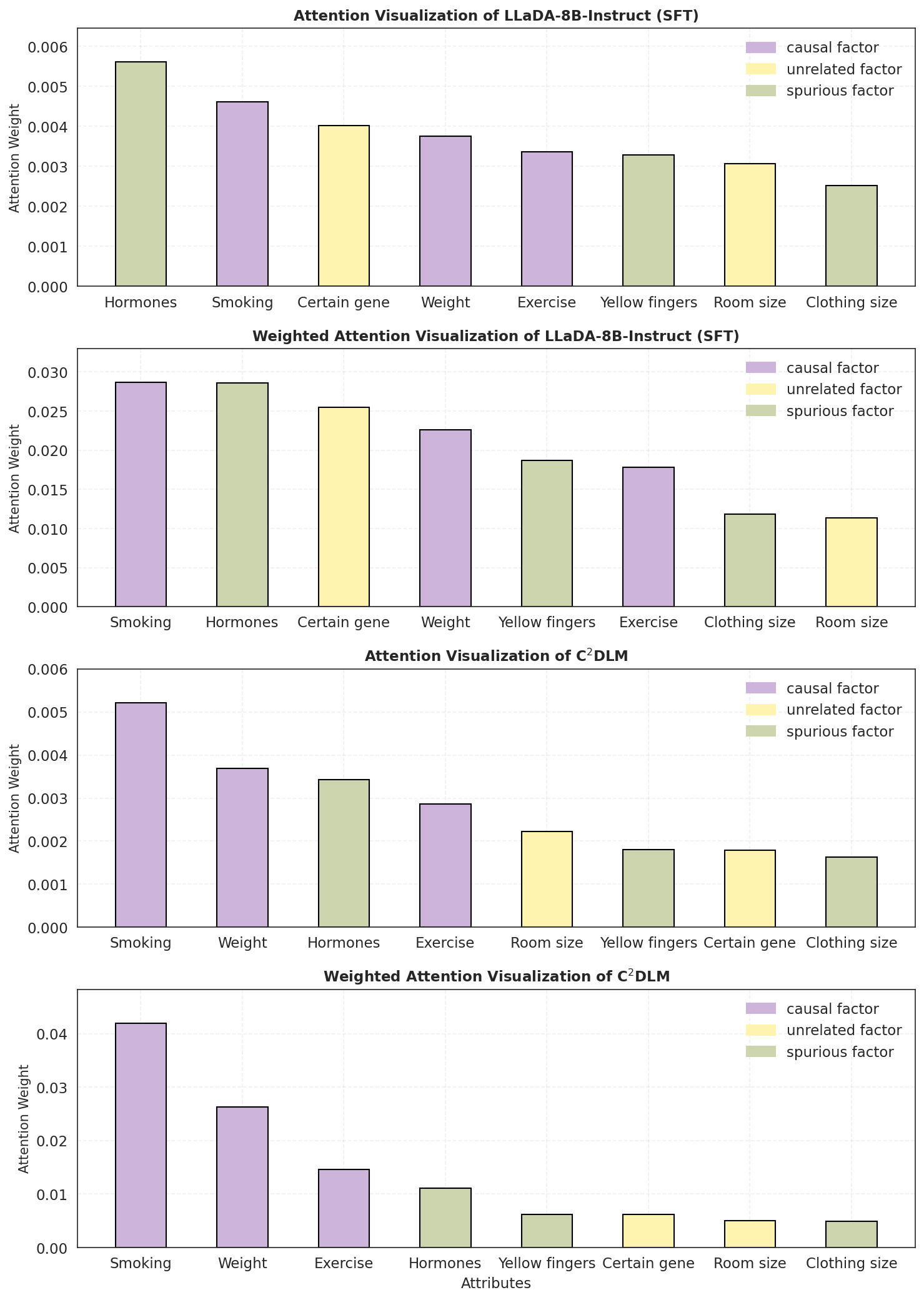}
    \caption{ Visualization of the weight distribution bar charts, presented from top to bottom: direct visualization of LLaDA’s attention weights bar chart; LLaDA’s attention weights weighted by the Value norm; direct visualization of C$^2$DLM’s attention weights bar chart; and C$^2$DLM’s attention weights weighted by the Value norm.}
    \label{fig:app_keshihua2}
\end{figure*}

This section provides a supplementary analysis of the attention visualizations presented in the main text. Specifically, the attention map illustrates how tokens interact within the model. However, examining the attention map alone is insufficient. For instance, in the phenomenon of attention sink, a disproportionate amount of attention is assigned to semantically insignificant tokens such as punctuation or prepositions. As compensation, the Value matrix of these tokens exhibit substantially lower norms compared to normal tokens. Consequently, using value-weighted attention offers a more accurate visualization target. Our visualization results are shown in Figure \ref{fig:app_keshihua1} and \ref{fig:app_keshihua2}. Based on the visualization, the following conclusions can be drawn:

Regardless of whether value weighting is applied, direct fine-tuning of LLaDA fails to effectively differentiate among the three types of factors, leading to a substantial decline in OOD performance. This occurs because the model primarily captures token correlations; when such correlations are disrupted in OOD settings, performance deteriorates significantly.

In contrast, C$^2$DLM effectively learns the underlying causal mechanisms of the model. As illustrated, the causal factors consistently exhibit greater importance than other factors, irrespective of whether value weighting is applied. Notably, although the Exercise attribute receives lower attention scores than Hormones in the raw attention map, its importance surpasses Hormones and ranks third once value weighting is considered. This highlights that weighted attention better captures the influence of value matrix norms across tokens, thereby providing a more faithful basis for analyzing token interactions.

\section{Human Evaluation of Teacher Model Generated Causal Graphs} \label{APP:Human}

Ideally, human experts should be employed for annotation. In this work, however, we use a teacher model to reduce costs and facilitate potential scaling. Accordingly, we conducted a human evaluation to verify the accuracy of the generated causal graphs. The evaluation procedure is detailed as follows:

We randomly sampled 50 annotated instances from the dataset, representing 7.3\% of the total data. Two human experts with undergraduate degrees in science and engineering independently conducted the evaluation. During the evaluation, the experts were allowed to use any online resources to search and cross-check concepts to ensure the accuracy of their assessments.

For the causal graphs, given the absence of a pre-established ground truth, we focused on the causal logical consistency of each edge. Specifically, for each effect, we checked whether the list of causes extracted by the LLM could logically account for it. If the causal relationship was correct, it was scored as 1; if the cause list was partially correct or incomplete, it was scored as 0.5; if incorrect, it was scored as 0.

The evaluation metric is computed as follows. Let $E_i = \{(c_{i,j}, e_i)\}$ denote the set of causal pairs, where $c_{i,j}$ is a cause and $e_i$ is the corresponding effect for instance $i$, and let $\text{score}(c_{i,j}, e_i)$ be the score assigned to each pair as described above. Then the accuracy for instance $i$ is:

\[
Acc_i = \frac{1}{|E_i|} \sum_{(c_{i,j}, e_i) \in E_i} \text{score}(c_{i,j}, e_i)
\]

The overall causal accuracy across all $N$ evaluated instances is:

\[
\text{Overall Accuracy} = \frac{1}{N} \sum_{i=1}^{N} Acc_i
\]

The experimental results are shown in Table \ref{Tab:human}, and can be summarized as follows:

\begin{enumerate}
    \item Two instances failed to generate causal graphs due to decoding errors, accounting for 4\% of the sampled data.
    \item For instances with correctly decoded causal graphs, the accuracy is 93.42\% $\pm$ 1.41\%.
\end{enumerate}

\begin{table*}[h!]
\centering
\begin{tabular}{lccccc}
\hline
Evaluator & Total Pairs & \#Score=1 & \#Score=0.5 & \#Score=0 & Average Accuracy \\
\hline
Human 1 & 311 & 280 & 28 & 3 & 94.84\% \\
Human 2 & 311 & 271 & 29 & 11 & 92.02\% \\
\hline
\end{tabular}
\caption{Evaluation results for human assessment of causal graphs.}\label{Tab:human}
\end{table*}

\section{Use Of AI Assistants}
We used generative AI, ChatGPT, to check for syntactic and grammatical errors in the manuscript. We carefully verified the correctness of the revised content.

\end{document}